\documentclass{article}

\usepackage{hyperref}
\usepackage{microtype}
\usepackage{graphicx}

\usepackage{booktabs} 
\usepackage[ruled,linesnumbered]{algorithm2e}
\usepackage{hyperref}
\usepackage{subcaption}
\usepackage{CJKutf8}
\usepackage{lineno}

\usepackage{multirow}
\usepackage[table]{xcolor}
\usepackage[accepted]{icml2025}

\usepackage{amsmath}
\usepackage[most]{tcolorbox}
\tcbuselibrary{breakable}
\usepackage{tcolorbox}
\usepackage{amssymb}
\usepackage{adjustbox}
\usepackage{mathtools}
\usepackage{colortbl}
\usepackage{amsthm}
\usepackage{makecell}
\usepackage[capitalize,noabbrev]{cleveref}
\usepackage{booktabs}
\theoremstyle{plain}

\theoremstyle{definition}

\theoremstyle{remark}

\usepackage{pifont}
\usepackage{tikz}
\usepackage[textsize=tiny]{todonotes}
\usepackage{listings}

\usepackage{xcolor}
\usepackage{soul}  
\sethlcolor{yellow!30}  

\usepackage[most]{tcolorbox} 
\newcommand{\roundedhl}[1]{%
    \tcbox[
        on line,
        valign=center,
        colback=gray!20,      
        colframe=gray!20,     
        boxrule=0pt,          
        boxsep=0pt,           
        left=1.5pt, right=1.5pt,
        top=1.5pt, bottom=1.5pt,
        arc=1.5pt,            
    ]{#1}%
}
\icmltitlerunning{Not All Documents Are What You Need for Extracting Instruction Tuning Data}

\begin{document}

\twocolumn[
\icmltitle{Not All Documents Are What You Need  for \\ Extracting  Instruction Tuning Data}

\icmlsetsymbol{equal}{*}


\begin{icmlauthorlist}
\icmlauthor{Chi Zhang}{equal,bit}
\icmlauthor{Huaping Zhong}{equal,sensetime}
\icmlauthor{Hongtao Li}{sensetime}
\icmlauthor{Chengliang Chai}{bit}
\icmlauthor{Jiawei Hong}{sensetime}
\icmlauthor{Yuhao Deng}{bit}
\icmlauthor{Jiacheng Wang}{bit}
\icmlauthor{Tian Tan}{ua}
\icmlauthor{Yizhou Yan}{meta}
\icmlauthor{Jiantao Qiu}{ailab}
\icmlauthor{Ye Yuan}{bit}
\icmlauthor{Guoren Wang}{bit}
\icmlauthor{Conghui He}{ailab}
\icmlauthor{Lei Cao}{ua}
\end{icmlauthorlist}

\icmlaffiliation{bit}{Department of Computer Science, Beijing Institute of Technology, Beijing, China}
\icmlaffiliation{sensetime}{Sensetime Research, Shenzhen, China}
\icmlaffiliation{ailab}{Shanghai Artificial Intelligence Laboratory, Shanghai, China}
\icmlaffiliation{ua}{University of Arizona, USA}
\icmlaffiliation{meta}{Meta, USA}
\icmlcorrespondingauthor{Chengliang Chai}{ccl@bit.edu.cn}
\icmlcorrespondingauthor{Conghui He}{heconghui@pjlab.org.cn}

\icmlkeywords{Machine Learning, ICML}

\vskip 0.3in
]

\printAffiliationsAndNotice{\icmlEqualContribution} 

\begin{abstract}
 Instruction tuning improves the LLMs performance but depends on high-quality training data. Recently, LLMs have been used to synthesize data, enhancing training with seeds like question-answer (QA) pairs. However, this synthesis often results in instruction examples similar to the seeds, lacking diversity and biasing real applications. Thus, we propose to extract instruction tuning data from web corpus with much rich knowledge. The most straightforward strategy is to quickly retrieve domain specific documents from the corpus and then extract all QA pairs of these documents for tuning LLMs, which has two main limitations. (1) Extracting all QA pairs using LLMs is prohibitively expensive; and (2) These extracted pairs are not all relevant with the downstream applications, and incorporating all of them for tuning may even hurt the model performance. To overcome the limitations,  we introduce \texttt{EQUAL}, an \textbf{E}ffective and scalable data extraction framework that iteratively interleaves  document selection and extract high-\textbf{QUAL}ity QA pairs to optimize instruction tuning. \texttt{EQUAL} first clusters the document set based on the embeddings generated by contrastive learning. Then it leverages the multi-armed bandit based strategy to quickly identify document clusters where can extract high-quality QA pairs for training. This iterative framework significantly reduces computational costs while improving model performance much. Experiments on AutoMathText and StackOverflow across four downstream tasks demonstrate that EQUAL reduces computational costs by 5–10$\times$ while improving accuracy by 2.5\% on LLaMA-3.1-8B and Mistral-7B. Code and data is available at \url{https://anonymous.4open.science/r/EQUAL-DC36/}.

\end{abstract}
\vspace{-2mm}
\section{Introduction}
Previous studies have shown that instruction tuning enables the powerful reasoning capability of Large Language Models (LLMs)~\cite{ouyang2022training, achiam2023gpt, dubey2024llama}, but requires sufficient high-quality training data~\cite{ntoutsi2020bias, yu2023metamath, shah2024ai}. However, although the weights of the open source LLMs are publicly available, the datasets employed to fine-tune these models are generally private. This lack of data accessibility limits the opportunities to effectively adapt Large Language Models (LLMs) to particular domains~\cite{Cobbe_Kosaraju_Bavarian_Hilton_Nakano_Hesse_Schulman_2021, hendrycks2021measuring}.

Recently, leveraging LLMs to synthesize data~\cite{li2024synthetic, yue2024mammoth2, luo2023wizardmath, yu2023metamath, li2024synthetic, ding2024unleashing} has attracted much attention as an effective solution to enrich the original training data based on some seeds ($e.g.$, original question-answer pairs, knowledge bases, etc.), thanks to the powerful understanding and generative capabilities of LLMs. However, the synthesized instruction data is hard to achieve high quality because
the LLMs-based generation tends to generate instruction examples similar to the seeds, lacking of diversity~\cite{guo2024human,li2024forewarned, li2024synthetic, xu2024magpie, ding2024unleashing} and deviating from the distribution of the downstream tasks. 

\noindent \textbf{Data Extraction from Documents.}
In reality, there is plenty of high-quality web corpus ($e.g.$, Common Crawl) which contains rich knowledge and can be leveraged as high-quality instruction data. 
However, this wealth of knowledge is widely spread within the corpus.
Recently, \citet{yue2024mammoth2} proposed a method to retrieve domain-specific documents from a large web corpus, followed by employing high-performance LLMs to extract QA pairs from these documents and then using the extracted QA pairs to fine-tune an LLM. However, it has the following limitations.
\begin{figure}[ht]
\vskip 0.2in
\begin{center}
\centerline{\includegraphics[width=\columnwidth]{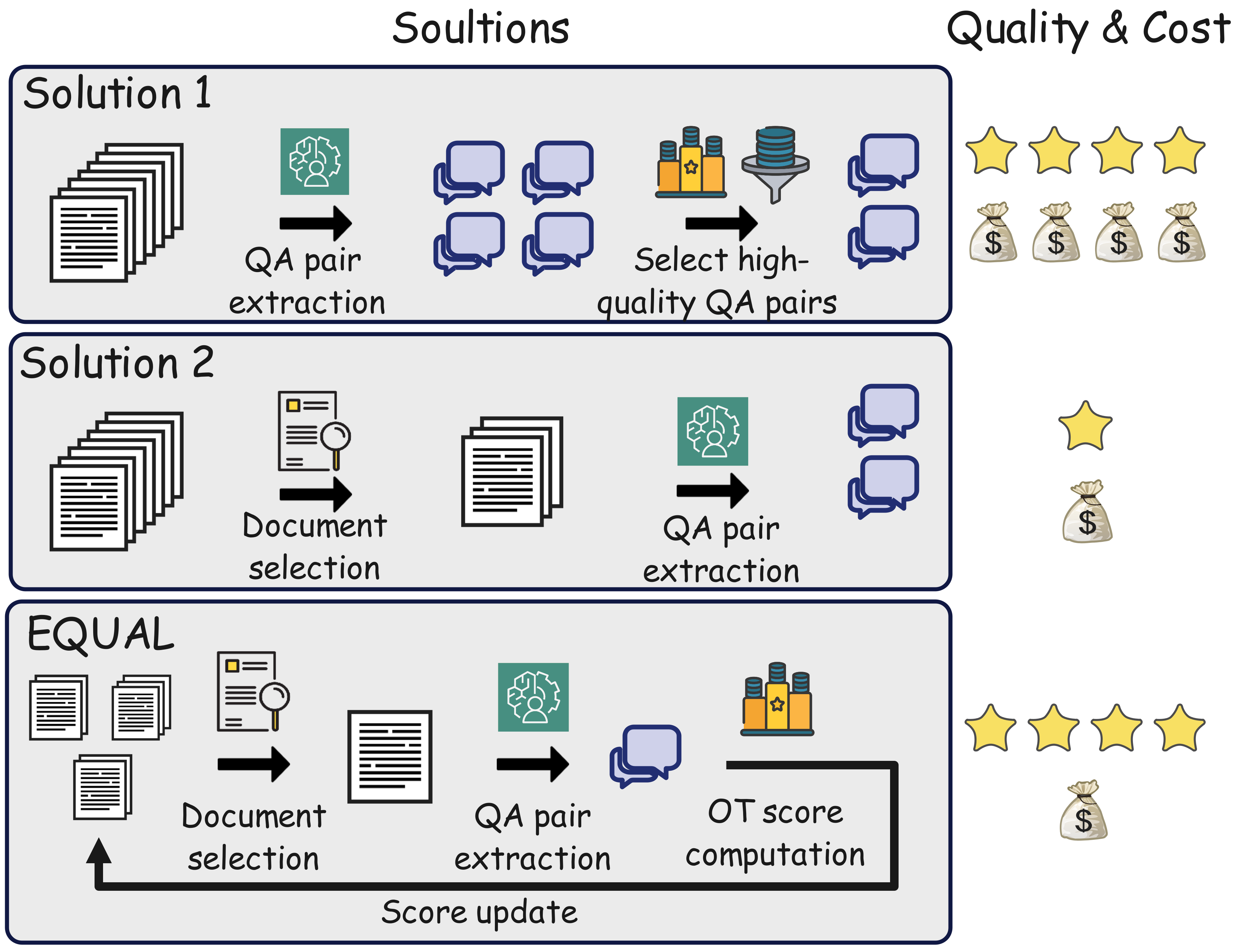}}
\vspace{-1em}
\caption{Comparison of Different Solutions.}
\label{fig:intro}
\end{center}
\vspace{-3em}
\end{figure}

\noindent \underline{\textit{Prohibitive Computational Cost.}}
To extract a high-quality instruction tuning dataset, it uses LLMs to repeatedly scan and analyze all the documents to extract question-answer (QA) pairs, each of which often requires multiple LLM calls~\cite{gilardi2023chatgpt, yue2024mammoth2}. Consequently, this process is prohibitively expensive, especially when there are a large number of documents to process. Solving this problem requires largely reducing the number of candidate documents, $e.g.$, by discovering the documents most valuable to instruction tuning. 

\noindent \underline{\textit{Irrelevant Instruction QA Pairs.}}
Even if an organization could afford extracting all domain-specific QA pairs from a large number of documents, blindly incorporating all of them to fine-tune an LLM could potentially degrade the model performance due to the presence of a significant amount of irrelevant data deviated from the target distribution~\cite{xialess, lin2024rho, xie2023data}.  Specifically,  a large corpus inevitably includes wild and noisy data with heterogeneous distribution that might differ significantly from the distribution desired by the downstream tasks. Therefore, it is necessary to judiciously identify high-quality QA pairs for extraction that are highly beneficial for downstream tasks.

\noindent \textbf{Intuitive Solutions.} There are two intuitive solutions to address the above limitations. Solution \raisebox{.7pt}{\small \textcircled{\hspace{-0.08cm} \raisebox{-.7pt}{1}}}: extract all QA pairs first and then 
select high-quality pairs.  
Solution \raisebox{.7pt}{\small \textcircled{\hspace{-0.08cm} \raisebox{-.7pt}{2}}}: Select high-quality documents that potentially contain high-quality pairs first and then extract QA pairs from them. Unfortunately, neither of the above two methods can solve both limitations. To be specific, for solution  \raisebox{.7pt}{\small \textcircled{\hspace{-0.08cm} \raisebox{-.7pt}{1}}}, although it can achieve good model performance, extracting all pairs beforehand is still costly.
Solution \raisebox{.7pt}{\small \textcircled{\hspace{-0.08cm} \raisebox{-.7pt}{2}}} is cost-effective, but it is difficult to accurately discover high-quality documents because documents and pairs have different feature distributions. Note that in this case, the quality of these documents should be measured by the potential contribution of the QA pairs to the target distribution, which is not aligned with the data quality of the original documents, e.g., dirty data, duplication. In other words, even if the embedding of a QA pair is close to that of a document, it does not necessarily indicate that the QA pair is close to the pair potentially extracted from the document. 


\noindent \textbf{Key Idea.} To address both limitations, our key idea is to interleave document selection and QA pairs extraction. During this iterative process, the extracted QA pairs help capture the relationship between the document and the pair distribution more and more accurately, and at the same time, the selected documents improve consistently.


\noindent \textbf{Our Proposal.} Inspired by the above idea, we propose \texttt{EQUAL}, a scalable and effective data extraction framework for constructing QA pairs from documents, aiming to enhance the LLMs instruction tuning. 
To be specific, \texttt{EQUAL} first clusters over the heterogeneous document set considering the feature similarities of QA pairs extracted from these documents.
To achieve this, we introduce a warm-up step using contrastive learning to align the feature space between documents and QA pairs.
In this way, \texttt{EQUAL} effectively identifies those high-quality clusters by sampling and extracting QA pairs from them to save cost.
Afterwards, we propose a Multi-arm Bandit (MAB) based technique to iteratively select the clusters. As the reward function, it predicts the benefit of QA pairs that potentially could be extracted from the clusters. More specifically, in each iteration, \texttt{EQUAL} tends to select the cluster where documents can produce QA pairs that are likely to benefit the target model performance. 
This benefit is measured by the \textit{optimal transport (OT) score}, where a higher benefit score indicates a smaller difference between the distributions of the QA pairs in the cluster and the target distribution. 
Then, given the selected cluster, \texttt{EQUAL} samples some documents from it, extracts QA pairs using LLMs, and in turn updates the optimal transport score of this cluster accordingly. In this iterative process, we precisely estimate the distribution of the QA pairs in a document cluster without having to conduct extraction over all documents. 

Moreover, leveraging the upper confidence bound technique in MAB, \texttt{EQUAL} promotes the potentially low-quality, thus under-sampled clusters. Therefore, it improves the diversity of the extraction data. This balance between exploration and exploitation effectively avoids reaching a local optimum.

To summarize, we make the following contributions:

\noindent (1) We propose \texttt{EQUAL}, a novel framework for data extraction from documents to enhance LLMs instruction tuning with high scalability.

\noindent (2)  We incorporate an iterative MAB solution to first cluster the documents and extract data from these clusters, achieving a good exploration-exploitation trade-off.

\noindent (3)  We propose a warm-up strategy to align the features of documents and QA pairs using contrastive learning.

\noindent (4) Extensive experiments on  datasets (AutoMathText and StackOverflow) with more than 1 million documents and 4 popular downstream tasks demonstrate that \texttt{EQUAL} significantly outperforms baseline methods by saving 5-10$\times$ 
computation resources consumption while still improving 2.5\% in accuracy (train/test on Llama-3.1-8B and Mistral-7B model).
\section{Preliminary}
We define the problem in \S \ref{sec:defination}, followed by the details of extracting QA pairs from a document.
\vspace{-2mm}
\subsection{Problem Definition}
\vspace{-2mm}
\label{sec:defination}
In this paper, we study the problem of data extraction from a candidate document pool $\mathcal{D}_c$ to extract QA pairs for instruction tuning. Formally, given $\mathcal{D}_c$ and a reference set $\mathcal{D}_r$ for instruction tuning, the problem is to select a subset $\mathcal{D}_b \subset \mathcal{D}_c$ from which a set of QA pairs $\mathcal{Q}$ are extracted to fine-tune an LLM $M$, aiming to minimize the loss of the updated model $M'$ on the reference dataset $\mathcal{D}_r$.
\vspace{-2mm}
\subsection{QA Pair Extraction}
\vspace{-2mm}
\label{sec:extract}
For a given document, we leverage advanced LLMs to extract and refine QA pairs. More specifically, we employ Qwen2.5-72B~\cite{yang2024qwen2} to extract QA pairs within the document. By incorporating  examples within the prompt, we guide the model to focus on the desired QA pairs while filtering out markups, boilerplate, and other irrelevant content. Although Qwen2.5-72B demonstrates impressive capabilities, the extracted QA pairs still exhibit issues such as improper formatting, missing answers, or mismatched responses~\cite{li2023self,honovich2022unnatural,chen2023alpagasus}. To address these problems, we employ further refinement to these QA pairs using Qwen2.5-72B, which has demonstrated significant enhancement in the quality of the extracted pairs~\cite{xu2023wizardlm}. The prompts for extraction and refinement are provided in Appendix~\ref{prompt}.


Overall, the entire extraction process requires each entire document as input and repeatedly calling high-performance LLMs, which is rather expensive. Thus, in this paper, we focus on reducing the number of documents to be extracted to save the cost while keeping high model accuracy. However, how to improve the quality of each extracted pair within each document as discussed above is orthogonal to this work.

\section{Proposed Approach.}
\begin{figure*}[ht]
\vspace{-.2mm}
\begin{center}
\centerline{\includegraphics[width=\textwidth]{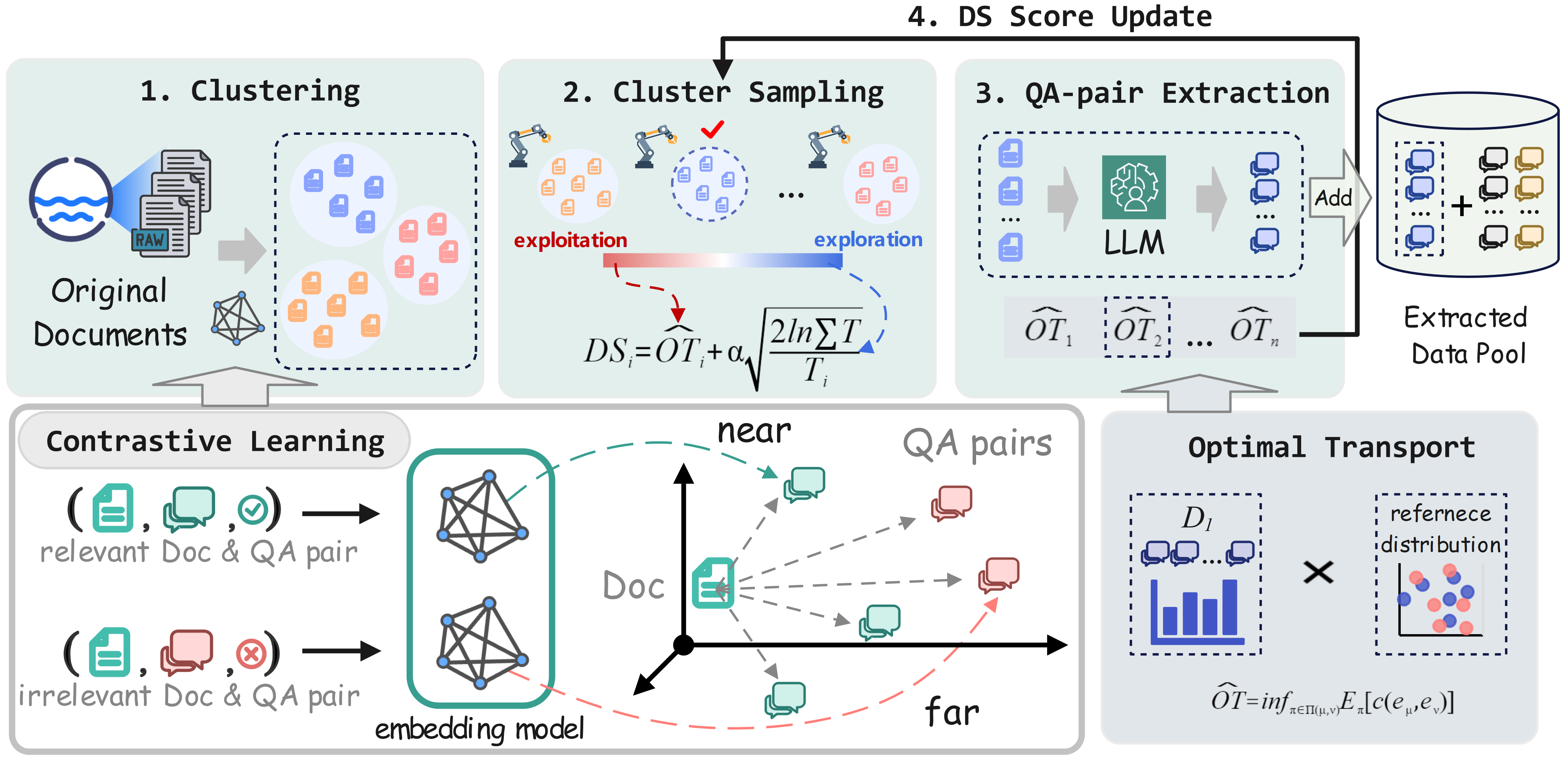}}
\caption{The Overall Framework of \texttt{EQUAL.}}
\label{fig:framework}
\end{center}
\vspace{-\baselineskip}
\vspace{-.2mm}
\end{figure*}
\subsection{The \texttt{EQUAL} Framework}
\label{sec:framework}
\textbf{Multi-Armed Bandit (MAB)}~\cite{vermorel2005multi} is an effective framework that makes decisions over time under uncertainty.
This consists of $N$ possible actions, each known as an $arm$. Pulling an arm indicates sampling from this arm to capture its reward distribution more accurately. This framework characterizes an agent that iteratively gains new knowledge by pulling arms that are rarely visited ($i.e.$, exploration) while using current knowledge to enhance its decisions by pulling arms already with a high reward ($i.e.$, exploitation).  The agent aims to balance the exploration and exploitation to maximize their overall reward throughout the given time span.

\textbf{Bridging \texttt{EQUAL} and MAB.}
The overall process of \texttt{EQUAL} is illustrated in Figure~\ref{fig:framework} and Algorithm~\ref{algorithm}. 
To reduce the computational cost,  we first cluster all documents in the candidate dataset $\mathcal{D}_c$ (line 1) such that the QA pairs extracted from each cluster are similar (Step-1 in Figure~\ref{fig:framework}, see  \S~\ref{sec:warmup} for details). Thus, we can neglect these low-quality clusters to save the cost. 
To precisely measure the quality of the cluster $C_i$, the most straightforward way is to extract all QA pairs and compare their distribution with that of the reference data, measured by the optimal transport score (see \S~\ref{sec:optimal_transport} for details), denoted by $OT_i$, but it is still very expensive. Hence, we propose to iteratively sample from these clusters to estimate the score.
Our key idea is inspired by a natural connection between \texttt{EQUAL} and MAB. 

At a high level, each cluster can be regarded as an arm of MAB. \texttt{EQUAL} iteratively selects a cluster, samples some documents and extract pairs from it ($i.e.$, pulling an arm). To be specific, as shown in Step-2 of Figure~\ref{fig:framework},  a cluster with a high estimated optimal transport score ($\hat{OT}_i$) tends to be selected.  The higher the score, the smaller difference between the distributions of QA pairs from this cluster and the reference data. 
Moreover, clusters that are rarely visited (denoted by the sampling frequency $T(C_i)$) tend to be selected as well to explore more diverse documents. 
Overall, putting $\hat{OT}_i$ and $T(C_i)$ together, we use the document sampling (DS) score to measure the quality of a cluster, which can achieve a good exploration-exploitation trade-off. 
Subsequently, as shown in Step-3, we obtain these extracted pairs (line 5-7) and update the score $\hat{OT}_i$ of the corresponding cluster (line 8). As more pairs from cluster $C_i$ are extracted, the estimated score $\hat{OT}_i$ will become more accurate.

%

Next, we illustrate the details of the \texttt{EQUAL} framework.

\textbf{DS Score Computation.} 
Following the upper confidence bound in the typical MAB framework, we define the DS score $DS_{j}$ of the cluster $C_j$ to effectively balance the exploration ($i.e.$, data diversity) and exploitation ($i.e.$, data quality) as follows.
\vspace{-\baselineskip}
\begin{equation}
    DS_{j} = \hat{OT_j}+\alpha\sqrt{\frac{2\ln {\sum_{C_k \in \mathcal{C}}{T(C_k)}}}{T(C_j)}}
\end{equation}
\noindent where $T(C_j)$ denotes the frequency of documents sampled from cluster $C_j$, $\sum_{C_k \in \mathcal{C}}{T(C_k)}$ denotes the total sampling times from all clusters.  $\alpha$  is set as $\frac{1}{\sum_{C_k \in \mathcal{C}} T(C_k) + 1}$~\cite{hao2019bootstrapping}, which provides higher weight to exploration in  early stages, but in later stages, it provides higher weight to exploitation (line 9-11). 

\textbf{Update the DS Score.} In each iteration, a subset of documents $B_i$ is sampled from the selected cluster $C_i$ with a high DS score, and a set of QA pairs $Q_i$ is then extracted from the documents in $B_i$. 
The OT score of $C_i$ will be updated as follows.
\begin{equation}
    \hat{OT_i} = \mathcal{OT}(\cup Q_i, \mathcal{D}_r)  , \quad T(C_i) += 1
\end{equation}
where $\cup Q_i$ denotes all the extracted QA pairs in cluster $C_i$ obtained from the beginning and $\mathcal{OT}(\cdot)$ denotes the function of computing the OT score.
Then we update the DS score of all clusters.
%

\textbf{Extracted Pairs Collection.} As shown in Figure~\ref{fig:framework}, in each iteration, we add the extracted QA pairs $Q_i$ to the extracted data pool $\mathcal{D}_e$. 
Finally, we  use the pairs in the pool to fine-tune the LLMs. 

\begin{algorithm}[h]
    \caption{\texttt{EQUAL} Algorithm}\label{algorithm}
    \KwIn{Candidate data pool $\mathcal{D}_c$, reference set $\mathcal{D}_r$, extracted data ratio $\gamma$.}
    \KwOut{Extracted data pool $\mathcal{D}_e$.}
    
    $\mathcal{C}$ = \texttt{Cluster}($\mathcal{D}_c$)\;
    
    $\mathcal{D}_e = \emptyset$\;
    
    \While{$\left|\mathcal{D}_e\right| < \gamma\left|\mathcal{D}_c\right|$}{
         Select cluster $C_{i}$ with the highest DS Score\;

        Sample $B_{i}$ documents  from $C_{i}$ \;

        Extract QA pairs $Q_{i}$ from $B_{i}$ \;

        $\mathcal{D}_e$ = $\mathcal{D}_e \cup Q_{i}$ \; 
                
        $\hat{OT_i} = \mathcal{OT}(\cup Q_i,\mathcal{D}_r),$ $T$($C_i$) += 1 \;  

        \For{$C_j$ in $\mathcal{C}$}{        
            $DS_{j}=\hat{OT_j}+\alpha\sqrt{\frac{2\ln {\sum_{C_k \in \mathcal{C}}{T(C_k)}}}{T(C_j)}}$ \;
        }
    }    \Return{$\mathcal{D}_e$}\;
\end{algorithm}
\vspace{-\baselineskip}
\subsection{Warm-up for Clustering}
\label{sec:warmup}

\noindent \textbf{Motivation.} Considering that the original documents contain much irrelevant content with downstream applications, there exists a discrepancy between the feature space of document embeddings and that of the QA pairs extracted from them. Obviously, we hope that similar QA pairs fall into the same cluster, but similar documents do not necessarily indicate similar pairs if we cluster purely based on  feature embeddings of documents. However, it is rather expensive to extract all pairs and then cluster.
Therefore, to improve the clustering quality, we propose to incorporate a warm-up step to align the two feature spaces using contrastive learning~\cite{khosla2020supervised}.


\noindent \textbf{Align via Contrastive Learning.} In the warm-up stage, we first randomly sample a small proportion of the documents from the candidate data pool and use LLMs to extract QA pairs. 
Then, we fine-tune the model ($i.e.$, \texttt{BAAI/bge-en-v1.5}) used for document embedding to capture the deep connection between the original documents and the extracted QA pairs. 
Specifically, we treat the sampled documents and these extracted QA pairs as positive training examples (denoted by \texttt{$(d, q^{+})$)} for contrastive learning. 
To generate negative examples $q^{-}$ for a document $d$, we conduct negative sampling from all current QA pairs.
Then we train with the following loss function:
\begin{equation}
    L = -log \frac{e^{sim(d,q^{+})}}{e^{sim(d,q^{+})} + \sum e^{sim(d,q^{-})}}
\end{equation}
where $sim$ denotes the cosine similarity between the embedding of a document $d$ and QA pair $q$. In this way, documents containing similar QA pairs tend to be closer in the embedding space and are thereby grouped together in the same cluster.
\vspace{-2mm}
\subsection{Optimal Transport Score}
\label{sec:optimal_transport}
\texttt{EQUAL} requires computing the distribution similarity of extracted data and reference data utilizing Optimal Transport (OT)~\cite{villani2009optimal}, which is widely adopted to compute the minimal cost of transforming one distribution into another.  Specifically, the lower the transportation cost, the closer the two distributions, indicating that the extracted data is more beneficial for the target distribution.
To be specific, suppose that the distribution of extraction data is $\mu$ and that of the reference set is $\nu$. The transportation cost from $\mu$ to $\nu$ can be calculated by $\mathcal{OT}(\mu,\nu)$ :
\begin{equation}
    \mathcal{OT}(\mu,\nu) \overset{\text{def}}{=} \inf_{\pi \in \Pi(\mu, \nu)} \mathbb{E}_{(e_\mu,e_\nu) \sim \pi} [c(e_\mu, e_\nu)]
\end{equation}
where $e_\mu$ and $e_\nu$ denote the embedding of extracted QA pairs $q_\mu,q_\nu$ from the two distributions, $\Pi(\mu, \nu)$ denotes the set of all joint distributions $\pi(e_\mu, e_\nu)$ with marginals $\mu(e_\mu)$ and $\nu(e_\nu)$. Here, $c(e_\mu, e_\nu) : \mathbb{X} \times \mathbb{X} \to \mathbb{R}$ is the cost function for moving $e_\mu$ to $e_\nu$, where $\mathbb{X}$ denotes the entire embedding space in \texttt{EQUAL}. To be specific, we use $1 - \frac{e_\mu^Te_\nu}{\|e_\mu\|\|e_\nu\|}$ as the transportation cost between $e_\mu, e_\nu$, which is a popular choice to measure the semantic dissimilarity~\cite{pennington2014glove}. 
Then given two distributions $\mu$ and $\nu$, there are numerous possible mappings ($i.e.$, $\pi \in \Pi$) between pairs from these distributions.  The cost for each mapping can be calculated by various $c(e_\mu, e_\nu)$ within the mapping, and the optimal transport (OT) score represents the minimum cost among all the mappings. 

Rather than evaluating the quality of each individual QA pair, optimal transport can capture the relationship between two distributions, which helps extract data better aligned with the target distribution.
\newcommand{\random}{\texttt{Random}}
\newcommand{\ifd}{\texttt{Perplexity}}
\newcommand{\influence}{\texttt{Influence}}
\newcommand{\embedding}{\texttt{Avg-sim}}
\newcommand{\equal}{\texttt{EQUAL}}
\newcommand{\all}{\texttt{All(Mammoth)}}
\newcommand{\qa}{\texttt{Rewriting}}
\vspace{-2mm}
\section{Experiment}
In this section, we fine-tune the base models in different domains and conduct sufficient ablation studies to demonstrate the efficiency and effectiveness of \equal.
\vspace{-2mm}
\subsection{Experiment Setup}
\begin{table*}[t]
\centering
\begin{adjustbox}{width=\textwidth}
\begin{tabular}{lc|ccc|ccc|ccc|ccc}
\toprule
\multirow{3}{*}{\raisebox{-\normalbaselineskip}}{\textbf{Model}} & & \multicolumn{6}{c}{\textbf{LLAMA-3-8B}} & \multicolumn{6}{c}{\textbf{Mistral-7B}}  \\
\cmidrule(lr){1-2} \cmidrule(lr){3-8} \cmidrule(lr){9-14}
{\textbf{Domain}} & & \multicolumn{3}{c}{\textbf{Math}} & \multicolumn{3}{c}{\textbf{Code}} & \multicolumn{3}{c}{\textbf{Math}} & \multicolumn{3}{c}{\textbf{Code}} \\
\cmidrule(lr){1-2} \cmidrule(lr){3-5} \cmidrule(lr){6-8} \cmidrule(lr){9-11} \cmidrule(lr){12-14}
\textbf{Methods} & \textbf{} & \textbf{GSM8K} & \textbf{MATH}  & \textbf{FLOPs}  & \textbf{HUMANEVAL} & \textbf{MBPP}  & \textbf{FLOPs} & \textbf{GSM8K} & \textbf{MATH} & \textbf{FLOPs}  & \textbf{HUMANEVAL} & \textbf{MBPP} & \textbf{FLOPs}  \\
\midrule
Base Model& & 55.19 & 23.04 & - & 31.1 & 51.9 & - & 45.10 & 14.80 & - & 23.2 & 41.8 & - \\
\midrule
Random &LoRA & 63.76& 30.26& \roundedhl{8.05}&31.1& 53.7&\roundedhl{6.32} &54.21&20.78&\roundedhl{7.96} &28.0 &45.0 &\roundedhl{6.06} \\
Avg-sim  &LoRA & 65.64& 30.12&\roundedhl{114.79} &31.7& 52.6&\roundedhl{65.53}  &51.33&21.86&\roundedhl{113.49} &28.1 &45.1 &\roundedhl{65.45} \\
Perplexity &LoRA & 63.61& 30.94&\roundedhl{134.35} &31.1 &54.6 &\roundedhl{71.07}&51.40 &22.18&\roundedhl{134.18} &25.6 &44.4 &\roundedhl{70.9} \\
Influnece &LoRA & 63.46& 28.10&\roundedhl{240.31} &33.5& 55.0&\roundedhl{130.46} &53.45 &18.62&\roundedhl{236.59} &26.2 &41.5 &\roundedhl{129.68} \\
Rewriting  &LoRA & 62.21& 27.05& \roundedhl{18.11}&30.7& 54.1& \roundedhl{13.15} &50.19&17.72& \roundedhl{17.90}& 26.3& 44.3&\roundedhl{13.01} \\
Perplexity-MAB &LoRA & 64.52& 30.48&\roundedhl{18.18} &34.6 &55.0 &\roundedhl{13.24} &51.78 &22.10&\roundedhl{17.36} &27.4 &45.6 &\roundedhl{12.99} \\
Influence-MAB &LoRA & 65.73& 30.78&\roundedhl{23.37} &35.3 &55.1 &\roundedhl{19.13} &55.12 &19.58&\roundedhl{18.44} &28.7 &44.7 &\roundedhl{18.18} \\
\midrule
\equal(Ours) &LoRA & \textbf{67.32}& \textbf{31.86}&\roundedhl{17.75} &\textbf{36.0} &\textbf{55.3} &\roundedhl{12.99} &\textbf{57.54}  &\textbf{23.56} & \roundedhl{17.57}& \textbf{31.3}& \textbf{46.7}&\roundedhl{12.64}  \\
\midrule
Random &FULL & 68.92&32.46&\roundedhl{8.83} &42.7& 52.3&\roundedhl{7.19}  &61.41 &25.76&\roundedhl{8.74} &31.1 &44.2&\roundedhl{6.84} \\
Avg-sim  &FULL & 70.35&33.18&\roundedhl{115.66} &46.1 &53.2 &\roundedhl{66.31}  &61.88 &26.16 &\roundedhl{114.44}&37.8 &45.1&\roundedhl{63.20}  \\
Perplexity &FULL & 64.52& 33.56&\roundedhl{150.28}&44.5 &50.5 & \roundedhl{71.85} &55.04 &27.38 &\roundedhl{148.03}&32.6 &44.7&\roundedhl{71.07} \\
Influence &FULL & 65.20& 29.64&\roundedhl{256.94} &39.6& 53.7&\roundedhl{131.24}  &56.18 &22.08&\roundedhl{248.97}&35.6 &45.6 &\roundedhl{129.51} \\
Rewriting  &FULL & 64.47& 30.62& \roundedhl{18.71}& 43.4& 50.6&  \roundedhl{13.78}& 57.21&22.67 & \roundedhl{18.33}& 33.3& 43.6&\roundedhl{13.67}  \\
Perplexity-MAB &FULL & 65.28& 32.92&\roundedhl{18.96}&44.5 &49.1 &\roundedhl{13.76}  &56.44&24.20 &\roundedhl{18.44}&34.1 &42.9 &\roundedhl{13.42} \\
Influence-MAB &FULL & 67.78& 32.86&\roundedhl{25.62}&46.3 &53.5 & \roundedhl{19.91} &60.42 &24.44&\roundedhl{19.30}&34.8 &44.0 &\roundedhl{18.79} \\
\midrule
\equal(Ours) &FULL & \textbf{73.01}& \textbf{35.10}&\roundedhl{18.55}&\textbf{49.4} &\textbf{56.3}  &\roundedhl{13.50}&\textbf{67.73} &\textbf{28.28} &\roundedhl{18.18}&\textbf{39.1} &\textbf{50.6}& \roundedhl{13.07}  \\
\midrule

\end{tabular}
\end{adjustbox}
\caption{Comparison with other algorithms in test accuracy (\%) on AutoMathText and StackOverflow. The best results are highlighted. We run each experiment for three times and report the average.}
\label{table1}
\vspace{-\baselineskip}
\end{table*}
\textbf{Training Settings.}
We evaluate \equal~using two foundational models ($i.e.$, LLAMA-3-8B and Mistral-7B) and two training settings, $i.e.$, full fine-tuning (FULL) and Low-Rank Adaption (LoRA). In both training scenarios, the batch size is set to 512 and the maximum learning rate is set as $1\times 10^{-5}$ with a cosine decay schedule.
For the FULL setting we train the extracted data for 2 epochs on 32 H100 GPUs, while for the LoRA setting we train the extracted data for 4 epochs on 16 H100 GPUs. For the warm-up stage, we randomly select 5\% documents from $\mathcal{D}_c$ to extract QA pairs and then use contrastive learning to fine-tune the original embedding model \texttt{BAAI/bge-en-v1.5 model}, and then it is employed to generate document embeddings   for subsequent clustering.

\textbf{Dataset Preparation.}
In our evaluation, we use  AutoMathText~\cite{zhang2024automathtext} and StackOverflow (created by us) datasets as the candidate data pool $\mathcal{D}_c$ for mathematical and coding tasks respectively. 
AutoMathText totally contains 4,936,292 documents, from which we select 1,459,288 ones by filtering documents with a metadata score of $lm\_q_1q_2\_score < 0.5$\footnote{$lm\_q_1q_2\_score$ is a metadata attribute for each document in AutoMathText ranging from [0, 1], which quantifies the document's relevance, quality, and educational value in the context of mathematical intelligence.}  to remove documents that are not very relevant about math.  We use the dataset after filtering as the candidate data pool. StackOverflow is crawled by us from \url{stackoverflow.com}, which contains 1,222,629 documents in total. Then we implement an $n$-gram filtering~\cite{guo2024deepseek} to ensure that our training data is not contaminated by information from the downstream tasks.

For the reference (validation) set $\mathcal{D}_r$, we respectively use the training set of GSM8K~\cite{cobbe2021training} and MBPP~\cite{austin2021program} for math and code domains, which are both widely used language modeling tasks and often serve as a validation benchmark for instruction tuning. 


During the clustering process, documents from $\mathcal{D}_c$ are clustered into 1,000 clusters using the $k$-means algorithm. The number of clusters is automatically determined by the Elbow~\cite{syakur2018integration} method in \equal. 

\textbf{Baselines.}
We compare \equal~with several baselines. 

(1) \random. We randomly sample documents to extract QA pairs from $\mathcal{D}_c$, which are then used for fine-tuning.

(2) \all. We fine-tune our model using the QA pairs extracted from all the documents in candidate data pool $\mathcal{D}_c$, which is the same method used in \texttt{Mammoth}~\cite{yue2024mammoth2}.

(3) \qa~\cite{yu2023metamath} synthesize new QA pairs based on existing QA pairs (specifically, the reference set $\mathcal{D}_r$ in our setting) using LLM. We synthesize the same number of pairs as other baselines.

(4) \embedding.  We extract QA pairs from all  documents in $\mathcal{D}_c$. Then we select QA pairs with the highest average similarities with the ones in $\mathcal{D}_r$. For each QA pair, we compute the embedding similarities between the pair and all pairs in $\mathcal{D}_r$, and compute the average.



(5) \ifd~\cite{li2024superfiltering} extract QA pairs from  all  documents in $\mathcal{D}_c$. Then we select  QA pairs with the highest perplexity scores. 

(6) \influence~\cite{xialess} extract QA pairs from  all  documents in $\mathcal{D}_c$. Then we select extracted QA pairs with the highest influence scores. 


(7) \ifd-\texttt{MAB} utilizes perplexity~\cite{li2024superfiltering} score as the reward of MAB to select documents for extracting QA pairs. 

(8) \influence-\texttt{MAB} utilizes influence~\cite{xialess} score as the reward, we select documents for extraction.

(9) \texttt{EQUAL} is our full-fledged solution.

\textbf{Metric.}
We evaluate the quality of extracted data by accessing the LLM performance, which has been fine-tuned with these data on several commonly used downstream tasks. (1) Math domain: GSM8K~\cite{cobbe2021training} and MATH~\cite{hendrycks2021measuring} are utilized for evaulating mathematical tasks. (2) Code domain: the fine-tuned LLM is evaluated on the HUMANEVAL~\cite{chen2021evaluating} and MBPP~\cite{austin2021program} datasets.  We also report FLOPs to quantify the total GPU cost across the following three stages: data extraction, data selection, and model training, with detailed information provided in Appendix~\ref{flops}.
\vspace{-5mm}
\subsection{Result}
\textbf{Overall Performance.}
In Table~\ref{table1},  we acquire 5\% documents for each baseline. We can observe that \equal~surpasses all the baseline methods on accuracy across all models and downstream tasks. 
%
%
Specifically, when implementing Full fine-tuning on Llama-3.1-8B, \equal~achieves an accuracy improvement of 4.09\% on GSM8k and 2.64\% on MATH compared with \influence, while saving approximately 5$\times$ $w.r.t.$ the computational cost.
\equal~surpasses \qa~due to the fact that the QA pairs generated by \qa~are quite similar to those QA pairs in $\mathcal{D}_r$, resulting in limited data diversity. Besides, the pairs directly generated by LLMs might be error-prone due to the hallucinations.
\equal~outperforms \ifd~and \ifd-\texttt{MAB} because perplexity score is solely based on the inherent complexity of the QA pairs to select extracted data without considering the downstream tasks. 
Besides, \equal~outperforms \influence~and \influence-\texttt{MAB} because influence function is easily affected by the length of the sequence~\cite{xialess}, often leading to the selection of pairs with fewer tokens.
Also, \equal~outperforms \embedding~because it selects data based on the similarities between QA pairs, without considering the overall distribution. 
%
In terms of the computational cost, we can observe that the FLOPs consumed by the \embedding, \influence~and \ifd~extraction method are notably high. This is due to their necessity of extracting the QA pairs from all the documents in $\mathcal{D}_c$, which incurs prohibitive cost. Additionally, since the influence score used in \influence~requires to compute gradients   during back propagation, leading to higher FLOPs consumption than  \equal. In contrast, it can be seen that for the influence score used in \influence~and the perplexity score used in \ifd, when combined with the MAB framework, comparable results can be achieved at a lower computational cost, which demonstrates the efficiency of the MAB strategy.

Table~\ref{table2} presents a comparison between \equal~and \random~using various QA pair extraction ratios. Interestingly, we find that the extracting of just 5\% of QA pairs for most tasks produces superior results compared to the use of complete $\mathcal{D}_c$. This demonstrates the effectiveness of \equal. Even for the difficult task MATH, extracting QA pairs from only 20\% to the documents in $\mathcal{D}_c$ can achieve comparable performance to \all~across all the training settings. This is because not all the QA pairs extracted from all the documents in $\mathcal{D}_c$ might contribute to the target tasks.
\vspace{-2mm}
\subsection{Ablation Study}
In this section, we demonstrate the effectiveness of Contrastive Learning (CL), Multi-Armed Bandit (MAB) and Optimal Transport (OT) through experiments conducted with \texttt{no-CL}, \texttt{no-MAB} and \texttt{no-OT} settings, which is shown in Table~\ref{table3}. Also, we conduct several ablation studies $w.r.t.$  the number of clusters, different clustering algorithms etc., and the results are illustrated in Figure~\ref{fig:ablation}.

\begin{table}[t]
\centering
\begin{adjustbox}{width=\columnwidth}
\begin{tabular}{lc|cc|cc|}
\toprule
\multirow{2}{*}{\raisebox{-\normalbaselineskip}}{\textbf{Domain}} & & \multicolumn{2}{c}{\textbf{\texttt{Math}}} & \multicolumn{2}{c}{\textbf{\texttt{Code}}}  \\
 \cmidrule(lr){1-2} \cmidrule(lr){3-4} \cmidrule(lr){5-6}
\textbf{Method} & \textbf{} & \textbf{GSM8K} & \textbf{MATH}  &  \textbf{HUMANEVAL} & \textbf{MBPP}   \\
\midrule
\random (5\%) &LoRA & 63.76& 30.26& 31.1& 53.7  \\
\random (10\%)  &LoRA & 66.03& 30.82&32.3&53.8  \\
\random (20\%)  &LoRA & 65.05& 31.76&33.5&54.1  \\
\all  &LoRA & 65.43&\underline{32.90} &34.8& 55.3  \\
\equal (5\%) &LoRA & 67.32& 31.86&36.0& 55.3  \\
\equal (10\%)  &LoRA & \underline{68.10}& 31.66&\underline{38.8}&\textbf{56.0}  \\
\equal (20\%)  &LoRA & \textbf{68.69}& \textbf{33.43}&\textbf{39.6}& \underline{55.5} \\
\midrule
\random (5\%) &Full & 67.40& 32.46&42.7&52.3   \\
\random (10\%)  &Full & 68.92& 34.54&43.6&54.6  \\
\random (20\%)  &Full & 70.05& 36.18&44.1&55.0  \\
\all  &Full & 70.28& \underline{40.02} & 45.6& 56.0   \\
\equal (5\%) &Full & 73.01& 35.10&49.4& \underline{56.3}  \\
\equal (10\%)  &Full & \textbf{74.46}& 38.19&\textbf{50.1}& 56.0 \\
\equal (20\%)  &Full & \underline{74.40}& \textbf{41.40}&\underline{49.6}& \textbf{56.4} \\
\midrule
\end{tabular}
\end{adjustbox}
\caption{Comparison with \random~at different ratios. The best results are highlighted in bold while the second-best results are underlined.}
\label{table2}
\vspace{-\baselineskip}
\end{table}
\textbf{Effectiveness of Contrastive Learning (CL).}
As shown in Figure~\ref{figure3}, points of the same color represent QA pairs extracted from the documents in the same cluster.  Figure~\ref{figure3-a} shows the clustering results obtained by directly using the existing model to compute embeddings for each document, while Figure~\ref{figure3-b} presents the results after fine-tuning the embedding model with CL. 
Since the original documents contain many contents irrelevant to the extracted QA pairs, leveraging the document embeddings  directly to cluster will lead to the inconsistencies of QA pair embeddings  within each cluster.
It can be observed that CL effectively aligns the embeddings of documents and their corresponding QA pairs. Also, we also conduct the experiment \texttt{no-CL}, which use the original embedding model \texttt{BAAI/bge-en-v1.5 model} to generate document embeddings for clustering. Following this, similar to \texttt{EQUAL}, both the MAB framework and optimal transport are employed to extract QA pairs iteratively based on the clusters. Table~\ref{table3} illustrates that \equal~surpasses \texttt{no-CL} across all experimental settings, demonstrating the effectiveness of CL.

\textbf{Effectiveness of MAB.}
In this section, we evaluate the performance of MAB  by contrasting it with a simple method called \texttt{no-MAB}, which extracts QA pairs directly from several clusters with distributions similar to the reference set without iterative extraction like MAB. Specifically, we start by sampling a small set of documents from each cluster and then extract QA pairs. Subsequently, optimal transport is employed to assess the similarity in distribution between these  pairs and the reference set, serving as the score for each cluster. Then, the documents within the top-scoring clusters are utilized to extract QA pairs and fine-tune the LLM. As illustrated in Table~\ref{table3}, \equal~outperforms \texttt{no-MAB}, because during the data extraction process, the OT score for each cluster is updated dynamically based on the previously extracted QA pairs, thereby enhancing the accuracy of the subsequent documents selection. 
\begin{figure}
  \begin{minipage}[t]{0.5\columnwidth}
    \centering
    \includegraphics[scale=0.075]{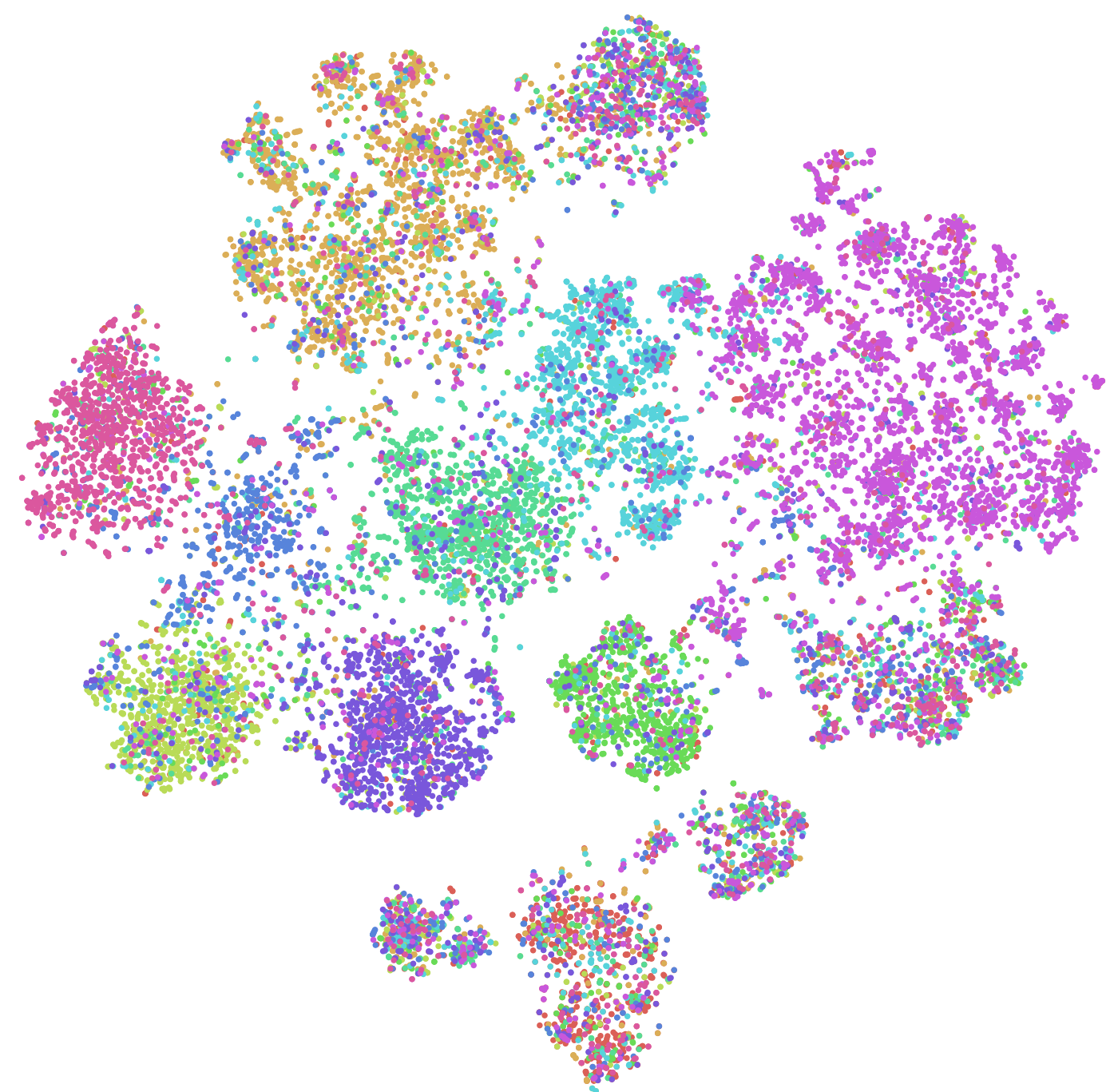}
    \subcaption{}
    \label{figure3-a}
  \end{minipage}%
  \begin{minipage}[t]{0.5\columnwidth}
    \centering
    \includegraphics[scale=0.11]{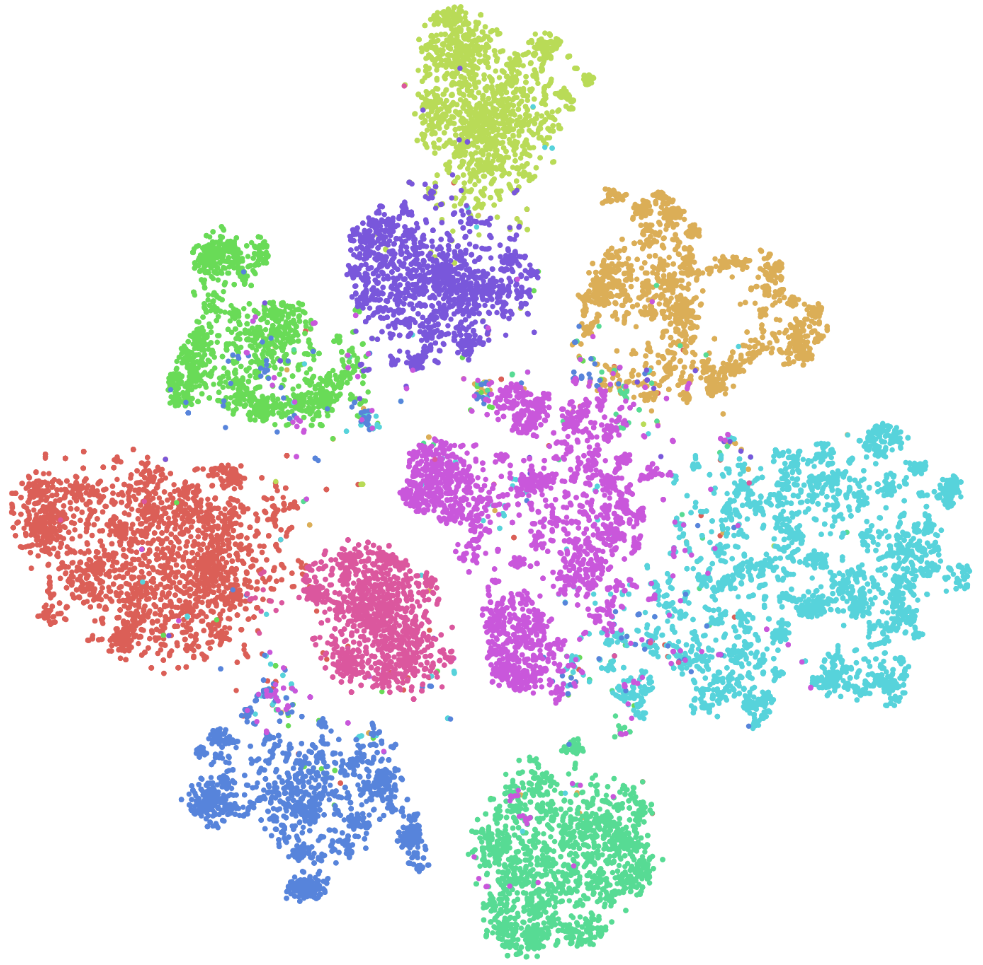}
    \subcaption{}
    \label{figure3-b}
  \end{minipage}%
  \caption{(a) shows the clusters based on the original embedding model; (b) shows the clusters based on the fine-tuned embedding model.}
  \label{figure3}
\end{figure}
\begin{table}[t]
\centering
\begin{adjustbox}{width=\columnwidth}
\begin{tabular}{lc|cc|cc|}
\toprule
\multirow{2}{*}{\raisebox{-\normalbaselineskip}}{\textbf{Domain}} & & \multicolumn{2}{c}{\textbf{\texttt{Math}}} & \multicolumn{2}{c}{\textbf{\texttt{Code}}}  \\
 \cmidrule(lr){1-2} \cmidrule(lr){3-4} \cmidrule(lr){5-6}
\textbf{Method} & \textbf{} & \textbf{GSM8K} & \textbf{MATH}  &  \textbf{HUMANEVAL} & \textbf{MBPP}   \\
\midrule
\texttt{no-CL} &LoRA & 64.05& 30.82& 32.8& 54.1 \\
\texttt{no-MAB}  &LoRA & 66.13& 30.55& 33.3& 53.3  \\
\texttt{no-OT}  &LoRA & 65.59& 31.08&34.4& 53.7 \\
\equal &LoRA & \textbf{67.32}& \textbf{31.86}&\textbf{36.0}& \textbf{55.3}  \\
\midrule
\texttt{no-CL} &Full & 69.73& 33.51& 44.3 &53.8   \\
\texttt{no-MAB}  &Full & 71.90& 33.25&46.9&55.5  \\
\texttt{no-OT}  &Full & 70.77 & 33.40&47.6 &54.6 \\
\equal&Full & \textbf{73.01}& \textbf{35.10}&\textbf{49.4}& \textbf{56.3}  \\
\midrule
\end{tabular}
\end{adjustbox}
\caption{Effectiveness of CL, MAB, OT in \equal.}
\label{table3}
\vspace{-\baselineskip}
\end{table}

\textbf{Effectiveness of Optimal Transport (OT).} We evaluate the effectiveness of optimal transport by comparing \equal~with \texttt{no-OT}, which utilizes the average similarities with QA pairs in $\mathcal{D}_r$ as the MAB reward. As shown in Table~\ref{table3}, the \equal surpasses the \texttt{no-OT} across all the settings. This indicates that optimal transport provides a more precise estimation of the distributional similarity between the extracted data and the reference set, hence better aligning the extracted data with downstream applications. 
\begin{figure}[ht]
\begin{center}
\centerline{\includegraphics[width=\columnwidth]{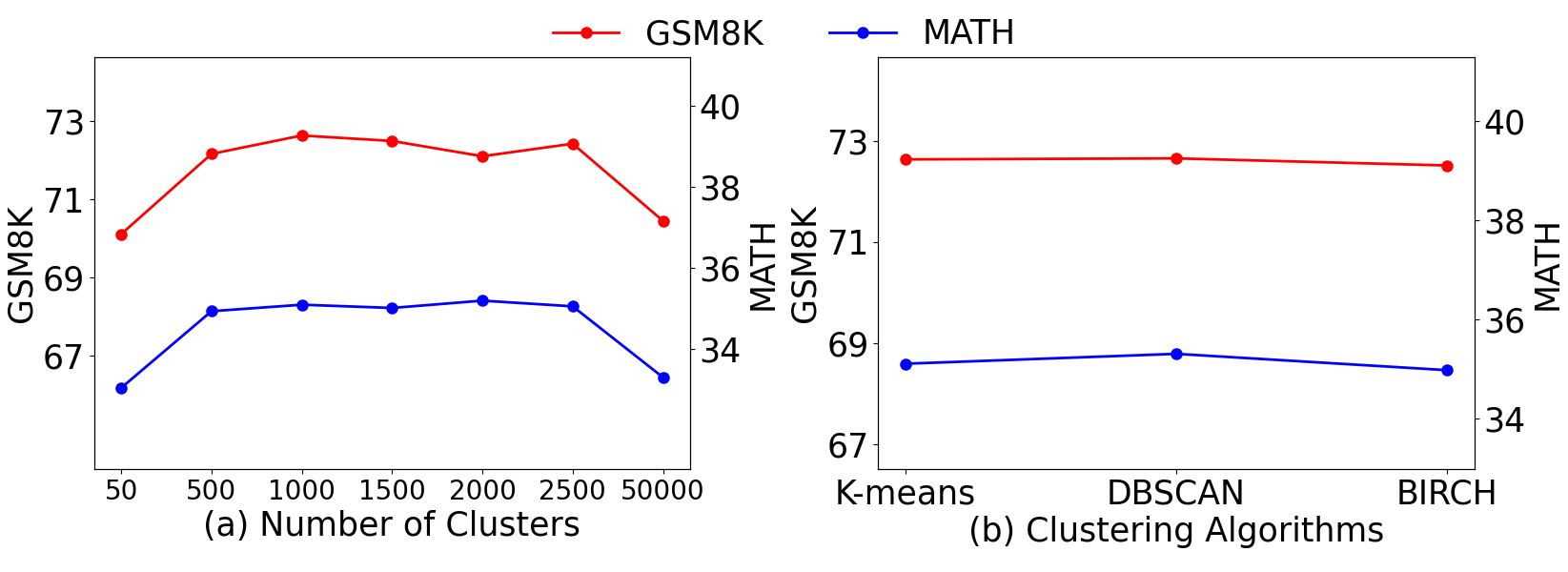}}
\caption{Ablation study of cluster numbers and clustering algorithms.}
\label{fig:ablation}
\end{center}
\vspace{-\baselineskip}
\end{figure}
\vspace{-\baselineskip}

\textbf{Number of Clusters.}
We use the Elbow~\cite{syakur2018integration} method to identify the optimal cluster numbers for  AutoMathText and StackOverFlow datasets.
In Figure~\ref{fig:ablation}(a), we plot the accuracy of \equal~with different cluster numbers. When the cluster number is around 1000 (the selected optimal number for both datasets), the model consistently performs well. However, a very small number of clusters ($i.e.$, $k=50$) leads to poor accuracy (2.90\% and 2.05\% lower  accuracy  on GSM8K and MATH tasks) due to the high variance of QA pairs extracted from the documents in each cluster. Thus, the QA pairs extracted from the sampled documents cannot well represent the cluster. Similarly, when the cluster number is too high~($i.e.$, $k=50,000$), there will be many clusters that are in fact contain similar QA pairs, and thus it is relatively hard to explore diverse clusters, thereby leading to the performance degradation  (2.56\% and 1.79\% lower accuracy).

\textbf{Clustering Algorithms.}
Moreover, we evaluate the performance of \equal~by other typical clustering algorithms including BIRCH~\citep{zhang1996birch} and DBSCAN~\citep{ester1996density}. The details of selecting optimal clustering parameters can be found in the Appendix~\ref{clustering}. As illustrated in Figure~\ref{fig:ablation}(b),  \equal~is robust to clustering algorithms on downstream tasks.
\section{Related Work}

Data synthesis and data selection are widely used techniques to enhance the LLMs performance during the instruction tuning phase.

\textbf{Data Synthesis using LLMs.} In data synthesis, high-performance LLMs are commonly used to generate complex and high-quality data~\cite{zhang2023self, yang2024self} training data.
For example, \citet{luo2023wizardmath, yu2023metamath} construct LLMs pipeline to revise existing training data, thus enhancing the quality and complexity. 
\citet{sun2024principle,wang2024step} conduct principle-driven prompting, which inserts some well-crafted principles into prompts to guide the LLMs for synthesis. 
\citet{guo2024human, li2024forewarned} iteratively synthesize important instruction tuning data and train the model with the newly synthesized data in each round. 
However, the data generated by these methods often exhibits low diversity, as few-shot prompts tend to make the newly generated data very similar to the original  data~\cite{li2024synthetic, ding2024unleashing}. 

To address this, researchers have proposed several techniques to generate diverse data. For instance, \citet{yu2024large, gupta2023targen} use various prompts to synthesize diverse data. \citet{yoo2021gpt3mix} integrates different existing training data to generate more diverse data.  \citet{divekar2024synthesizrr} uses retrieval augmentation to enhance data diversity by feeding LLMs different retrieved contents. Closer to our work, \citet{yue2024mammoth2} synthesizes QA pairs from a vast collection of web documents to enhance the diversity of synthetic data.


\textbf{Data Selection for Instruction Tuning.}
High-quality data plays a critical role in instruction tuning~\cite{brown2020language,zhou2024lima,xialess}. Simple approaches such as rule-based methods~\cite{soldaini2024dolma, penedo2023refinedweb} and deduplication~\cite{abbas2023semdedup} can improve data quality but the improvement is limited due to simple heuristics. 
More sophisticated methods like LLMs like GPT-4 assess data based on human-defined metrics but this is rather expensive~\cite{wettig2024qurating}. 
Perplexity-based methods~\cite{marion2023less, li2024superfiltering} select data samples that are difficult for the model to predict.
But all above methods do not consider the target distribution of downstream applications.
The influence function can measure the impact of training data on downstream model performance~\cite{grosse2023studying, xialess}, but is computationally intensive and affected by the length of the sequence~\cite{xialess}, leading to imprecise filtering. 
\citet{shao2024deepseekmath, yu2024mates} develop a surrogate model trained on available high-quality data to efficiently select data from the candidate dataset; however, its effectiveness is limited by the generalizability of the model.
%
Compared with the above methods, the optimal transport used in \texttt{EQUAL} efficiently measures the distribution difference between the selected data and reference data, which sufficiently reflects the downstream model performance. 

\section{Conclusion}
\vspace{-1mm}
This paper present \texttt{EQUAL}, a scalable and effective method for instruction tuning data extraction. \texttt{EQUAL} first use contrastive learning to unify the embedding feature spaces of the original documents and the extracted QA pairs. Based on this, \texttt{EQUAL} clusters all candidate document and regards each cluster as an arm of MAB framework due to uncertain distribution similarity scores, allowing sampling from quality clusters to estimate distribution similarity scores accurately while maintaining diversity.

\section*{Impact Statement}
This paper presents work whose goal is to advance the field of creating instruction tuning data.
We provide a new perspective of synthesizing training data from web corpus with rich knowledge, leading to  high-quality and diverse training data. With  novel techniques, we also achieve cost-effective extraction using LLMs.


\nocite{langley00}
\bibliography{example_paper}
\bibliographystyle{icml2025}
\newpage
\appendix
\onecolumn
\section{Ablation Study of Clustering Numbers And Algorithm}
\label{clustering}

\noindent[\textit{Metric and Criteria.}]
We use the metric Within-Cluster Sum of Squares (WCSS) to select the best cluster number using the well-known Elbow~\cite{syakur2018integration} algorithm. WCSS is the sum of squared distances between each data instance and its cluster center, i.e., WCSS=$\sum_{i=1}^k\sum_{x \in C_i}\|x-\mu_i \|$. At a high level, the criteria should be that within each cluster, data instances are close to each other, based on which it is better for different cluster centers to be far away from each other. Based on the criteria, the Elbow algorithm leverages the WCSS as a measurement to iteratively select an appropriate cluster number, as follows.

\noindent[\textit{Specific hypermarameter selection strategy.}]  
To be specific, Elbow begins with a small $k$, and with $k$ increasing, WCSS first decreases rapidly and then slows down.  Then, we identify the "elbow point" where the decreasing rate becomes slow as the best $k$. Thus, within each cluster, data points are sufficiently close to one another. Furthermore, given that $k$ remains modest, different cluster centers tend to maintain a distance from each other.

\noindent\underline{\textit{Clustering algorithms.}} In terms of the clustering algorithms, we also added experiments to show that \texttt{EQUAL} is not sensitive to clustering algorithms mainly because different algorithms have their own strategies to select appropriate parameters, which follows the criteria mentioned above. 

Specifically, we evaluate the performance of several typical clustering methods including BIRCH~\cite{zhang1996birch} and DBSCAN~\cite{ester1996density}.
Considering that the clustering results are easily affected by the parameters of clustering algorithms, we use different methods to select proper parameters. For DBSCAN, there are 2 key parameters: (1) $eps$(the radius of a neighborhood w.r.t. some data points) and (2) $minPts$ (a data point is considered as a core point if at least $minPts$ data points are within $eps$ of it). They can be set using the method in \citep{schubert2017dbscan}. For BIRCH~\citep{zhang1996birch}, we can use the Elbow~\cite{syakur2018integration} algorithm or Sihouette score~\cite{shahapure2020cluster} to determine the appropriate number of components.
\section{FLOPs Calculation}
\label{flops}
FLOPs is the number of floating point operations performed by GPUs. Many state-of-the-art methods [1,2,3] use it to measure the consumption of  GPU computing resources. In our experiments, FLOPs is collected directly in the data selection process using the Python code: \\
\textbf{\texttt{import}} \texttt{torch} \\
\textbf{\texttt{import}} \texttt{torch.nn \textbf{\texttt{as}} nn} \\
\textbf{\texttt{from}} \texttt{torch.profiler \textbf{\texttt{import}} profile, ProfilerActivity} \\ \\
\texttt{model = nn.Linear(1024, 512).cuda()} \\
\texttt{input\_data = torch.randn(128, 1024).cuda()} \\
\textbf{\texttt{with}} \texttt{profile(activities=[ProfilerActivity.CPU, ProfilerActivity.CUDA],} \\
\hspace*{2em} \texttt{with\_flops=True) \textbf{\texttt{as}} prof:} \\
\hspace*{4em} \texttt{model(input\_data)} \\
\textbf{\texttt{print}} \texttt{(prof.key\_averages().table(sort\_by="flops", row\_limit=10))}

\section{Length of Selected Data}
\begin{table}[h]
\centering
\begin{adjustbox}{width=0.5\textwidth}
\begin{tabular}{lcccc}
\midrule
\textbf{Length} & \textbf{Random} & \textbf{Influence} & \textbf{Perplexity}  &  \textbf{EQUAL(Ours)}   \\
\midrule
Prompt  &48.99 &37.86 &49.58 &58.38   \\
Output   &470.05 &222.94 &1235.90 &438.69 \\
Total   & 519.04& 260.80&1285.48& 497.07 \\
\midrule
\end{tabular}
\end{adjustbox}
\caption{The average length of extracted QA pairs with different methods ($i.e.$, \texttt{Influence}, \texttt{Perplexity} and \texttt{EQUAL(Ours)})}
\label{table7}
\end{table}

\section{Prompts for QA pair Extraction}
\label{prompt}
\subsection{Coding Task}
\begin{tcolorbox}[colback=gray!50!yellow!20,
    colframe=black!50!green!20,
    width=17cm,
    arc=2mm, auto outer arc,
    title={Code},breakable,]		
    \subsection*{System:}
    
    You are given a set of pre-processed documents, each of which may contain natural question-answer (Q-A) pairs. Your task is to identify and extract these pairs while ignoring unrelated content such as ads, markup, or boilerplate text.

    Input: 
    
    Each document contains multiple sections of text. Some of these sections may have clear questions followed by answers, while others may be irrelevant (e.g., ads or noise).

    Output: 
    
    Extract the Q-A pairs found within each document. A valid Q-A pair must consist of a clearly defined question and its corresponding answer. If no natural Q-A pair exists in the document, return void for that document. In the document, in order to describe the problem more clearly, the questioner usually attaches some useful information (e.g., code or explaination) to make it easier for others to better understand the problem. You need to extract this part of the content that needs to be complete as well.

    Here are some examples:
    
    \# Example 1
    
    Content: 
    
    Sorting lines date-wise and time-wise using Python from a \texttt{.txt} file. I have just written a Python code to extract data from around 700 text files into one file called \texttt{out\_data.txt}. The contents of the \texttt{out\_data.txt} file look something like this:

\begin{lstlisting}
datetime,V_1,V_2,V_3,V_4,V_5,V_6,V_7
2013-03-17 18:01:48.372,100,884,776,009,6553,ffff,987
2013-03-17 18:02:03.828,876,632,887,008,5423,879,443
2013-05-17 20:13:52.488,543,987,233,112,098,344,123
2013-08-17 23:09:08.171,667,9887,9897,09876,0987,098,0987
2013-01-17 35:06:04.172,267,987,6897,9876,1287,3498,2987
...
\end{lstlisting}

There are a total of 5,783,374 lines in the \texttt{out\_data.txt} file, and each line (after the header) begins with the datetime value.

However, the problem I have is that the code I wrote extracts the data from each individual file and adds it to my \texttt{out\_data.txt} file, but the lines are not in the order of date-time as you can see above. I was hoping to get my lines to be in date-time order because I need to plot this data. Any help will be highly appreciated !

Here is my current code:

\begin{lstlisting}
import re  \# Regular expressions
import glob  \# File management and reading

if __name__ == "__main__":  \# Opening for Python
    all_header = []  \# List declaration
    all_values = []  \# List declaration
    i = 0
    with open('out_data.txt', 'w') as of:  \# Output file
        for infile in glob.glob("/Users/name/Desktop/raw_data/*.txt"):  \# Input file
            with open(infile) as fobj:
                print(f"Processing file {infile}")
                for line in fobj:
                    data = line.split()  \# Split each line into individual tokens
                    if len(data) == 2 and re.search(r'(\d+-\d+-\d+)', data[0]):  \# Regular expression to identify date and time
                        header = ['datetime']  \# Column name datetime
                        values = [data[0] + " " + data[1]]  \# date+time as one value
                    else:
                        header = [d for d in data if data.index(d) % 2 == 0]
                        values = [d for d in data if data.index(d) % 2 != 0]
                    all_header.extend(header)
                    all_values.extend(values)
                    if not header:
                        if i == 0:
                            of.write(','.join(all_header))
                        i += 1
                        of.write("\n")
                        of.write(','.join(all_values))
                        all_header = []
                        all_values = []
        of.write("\n")
        of.write(','.join(all_values))
\end{lstlisting}

\subsection*{Expected Result}
The expected result from the example data given above would be:

\begin{lstlisting}
datetime,V_1,V_2,V_3,V_4,V_5,V_6,V_7
2013-01-17 35:06:04.172,267,987,6897,9876,1287,3498,2987
2013-03-17 18:01:48.372,100,884,776,009,6553,ffff,987
2013-03-17 18:02:03.828,876,632,887,008,5423,879,443
2013-05-17 20:13:52.488,543,987,233,112,098,344,123
2013-08-17 23:09:08.171,667,9887,9897,09876,0987,098,0987
\end{lstlisting}

However, I could not figure out how to include the sort element in the code or if there is any other way to achieve this.

\subsection*{Solution Using \texttt{pandas}}
You can use \texttt{pandas}. A simple example would be as follows:

\begin{lstlisting}{python}
import pandas as pd
import glob

df_list = []
for infile in glob.glob("/Users/name/Desktop/raw_data/*.txt"):
    df_list.append(pd.read_csv(infile, parse_dates=['datetime']))
df = pd.concat(df_list).sort_values(by='datetime')
df.to_csv('out_data.txt', index=False)
\end{lstlisting}

\subsection*{Solution Using \texttt{csv}}
An alternative method is:

\begin{lstlisting}{python}
import csv

with open("out_data.txt", "r") as f:
    reader = csv.reader(f, delimiter=",")
    header = next(reader)
    sortedlist = sorted(reader, key=lambda x: x[0])

with open("sorted.txt", "w") as f:
    writer = csv.writer(f, lineterminator="\n")
    writer.writerow(header)
    writer.writerows(sortedlist)
\end{lstlisting}

\subsection*{Solution Using Bash}
As an alternative, you can also use Bash:

\begin{lstlisting}[language=bash]
head -1 out_data.txt > sorted.txt
tail +2 out_data.txt | sort -t, -k1 >> sorted.txt
\end{lstlisting}

Hope this helps.

\section*{Q:}
I've just written a Python code to extract data from around 700 text files into one file called \texttt{out\_data.txt}. 

The contents of the \texttt{out\_data.txt} file look something like this:

\begin{lstlisting}
datetime,V_1,V_2,V_3,V_4,V_5,V_6,V_7
2013-03-17 18:01:48.372,100,884,776,009,6553,ffff,987
2013-03-17 18:02:03.828,876,632,887,008,5423,879,443
2013-05-17 20:13:52.488,543,987,233,112,098,344,123
2013-08-17 23:09:08.171,667,9887,9897,09876,0987,098,0987
2013-01-17 35:06:04.172,267,987,6897,9876,1287,3498,2987
...
\end{lstlisting}

There are a total of 5,783,374 lines in the \texttt{out\_data.txt} file, and each line (after the header) begins with the \texttt{datetime} value.

However, the problem I have is that the code I wrote extracts the data from each individual file and adds it to my \texttt{out\_data.txt} file, but the lines are not in the order of \texttt{datetime} as you can see above. I was hoping to get my lines to be in \texttt{datetime} order because I need to plot this data.

Any help will be highly appreciated.

\section*{A:}
\begin{lstlisting}{python}
import re  \# regular expressions
import glob  \# file management and reading

if __name__ == "__main__":  \# opening for Python
    all_header = []  \# list declaration
    all_values = []  \# list declaration
    i = 0
    with open('out_data.txt', 'w') as of:  \# output file
        for infile in glob.glob("/Users/name/Desktop/raw_data/*.txt"):  \# input files
            with open(infile) as fobj:
                print("processing file {}".format(infile))
                for line in fobj:
                    data = line.split()  \# split each line into individual tokens
                    if len(data) == 2 and re.search(r'(\d+-\d+-\d+)', data[0]):  \# identify date
                        header = ['datetime']  \# column name
                        values = [data[0] + " " + data[1]]  \# combine date and time
                    else:
                        header = [d for d in data if data.index(d) % 2 == 0]
                        values = [d for d in data if data.index(d) % 2 != 0]
                    all_header.extend(header)
                    all_values.extend(values)
                    if not header:
                        if i == 0:
                            of.write(','.join(all_header))
                        i += 1
                        of.write("\n")
                        of.write(','.join(all_values))
                        all_header = []
                        all_values = []
        of.write("\n")
        of.write(','.join(all_values))
\end{lstlisting}

\subsection*{Expected Result:}
From the example data given above, the output should be:

\begin{lstlisting}
datetime,V_1,V_2,V_3,V_4 ,V_5 ,V_6  ,V_7
2013-01-17 35:06:04.172,267 ,987 ,6897,9876,1287,3498 ,2987
2013-03-17 18:01:48.372,100 ,884 ,776 ,009 ,6553,ffff ,987
2013-03-17 18:02:03.828,876 ,632 ,887 ,008 ,5423,879  ,443
2013-05-17 20:13:52.488,543 ,987 ,233 ,112 ,098 ,344  ,123
2013-08-17 23:09:08.171,667 ,9887,9897,09876,0987,098  ,0987
\end{lstlisting}

\subsection*{Using Pandas:}
You can use the Pandas library for simplicity. Here is an example:

\begin{lstlisting}{python}
import pandas as pd
import glob

df_list = []
for infile in glob.glob("/Users/name/Desktop/raw_data/*.txt"):
    df_list.append(pd.read_csv(infile, parse_dates=['datetime']))
df = pd.concat(df_list).sort_values(by='datetime')
df.to_csv('out_data.txt', index=False)
% \end{lstlistings}

\subsection*{Using CSV Module:}
You can also perform an ordinary (dictionary order) sort as follows:

% \begin{lstlistings}{python}
import csv

with open("out_data.txt", "r") as f:
    reader = csv.reader(f, delimiter=",")
    header = next(reader)
    sortedlist = sorted(reader, key=lambda x: x[0])

with open("sorted.txt", "w") as f:
    writer = csv.writer(f, lineterminator="\n")
    writer.writerow(header)
    writer.writerows(sortedlist)
\end{lstlisting}

\subsection*{Using Bash:}
Alternatively, you can use the following Bash commands:

\begin{verbatim}
head -1 out_data.txt > sorted.txt
tail +2 out_data.txt | sort -t, -k1 >> sorted.txt
\end{verbatim}

\noindent Hope this helps!

\end{tcolorbox}

\subsection{Mathematical Task}

\begin{tcolorbox}[colback=gray!50!yellow!20,
    colframe=black!50!green!20,
    width=17cm,
    arc=2mm, auto outer arc,
    title={Math},
    breakable,]		
\textbf{System:}

You are an excellent AI assistant who is good at constructing question-answer (Q-A) pairs. Your task is to construct some math Q-A from the original documents.

Input:

Each document contains multiple sections of text. Some of these sections may contain mathematical content which can be used to construct Q-A pairs.

Output: 

Identify valid content and construct Q-A pairs. A valid Q-A pair must consist of a clearly defined question and its corresponding answer. Specially, the questions should be solvable that provide valid and complete pre-conditions; and the answers need to satisfy the Chain of Thought (CoT) format, which instructs the responder to solve the question step by step. If the content in the document is not suitable for Q-A construction, return void for that document.

Here is an example:

\textbf{User}: As I mentioned certain scientific terms in my previous post, I would like to go in-depth on those concepts, beginning with terminal velocity, it being the most fundamental concept in my post.

\textbf{So what is terminal velocity?}

Terminal velocity is the velocity of an object when the drag force (dependent on the fluid the object is travelling through) acting upon it is equal to the downward force of gravity acting upon it. Simply put, when the air resistance of a falling object cancels out the gravitational force which is pulling it downwards and accelerating it.

\textbf{So how do these forces affect the motion of the object?} The forces cancelling each other out make the object remain at a constant rate of motion.

You may ask why does the object still move when the forces cancel each other out. This is due to the fact that in the beginning the force of gravity still manages to overcome the drag force, allowing the object to gain speed (accelerate) initially. But as the object increases in velocity, the drag force increases. This effect can also be seen in the case of friction (Drag and friction are pretty much the same thing). Let's assume that a boy is dragging a heavy box, full of files, across a distance of 100 meters. Now, we will imagine this scenario in two different ways: firstly, in the case whereby the boy is walking slowly, and in the second, whereby the boy is running. So in the first case, the boy walks; when he reaches the end, he feels the bottom of the box, where the box and the floor meet, it still feels the same as before. Now in the second case, he runs; he once again feels the bottom of the box, this time it feels warmer than before. 

\textbf{So what can we infer from this scenario?}

Before I reveal the answer, I would like to state a few properties of friction:
\begin{itemize}
    \item Friction opposes motion
    \item Friction causes wear and tear
    \item Friction produces heat when kinetic energy is converted into thermal energy
\end{itemize}

\textbf{So what can we infer?} In the second scenario, there was more heat; therefore, we can assume that there was more frictional force produced in the second case.

Now let's go back to what I mentioned previously, air resistance increases (Drag Force) as the object’s velocity increases. As seen in the example above, we can tell that this statement is true.

\textbf{Recap}:
\begin{itemize}
    \item Terminal velocity is the velocity an object is at when the gravitational force acting upon it is equal to the drag force acting upon it in the opposite direction, therefore cancelling out all forces and resulting in a resultant force of 0.
    \item The drag force acting upon the object increases as the object accelerates due to the downward force of gravity.
\end{itemize}

Ok, so let's move on to the math behind terminal velocity and some examples of it.

The formula for terminal velocity is as follows:
\[
V_t = \sqrt{\frac{2mg}{\rho A C_d}}
\]

where:
\begin{itemize}
    \item \( m \) = Mass of falling object
    \item \( g \) = Acceleration of the object due to gravity
    \item \( \rho \) = Density of fluid through which the object is travelling
    \item \( A \) = Projected area of the object
    \item \( C_d \) = Drag Coefficient
\end{itemize}

Example:

Assuming I drop a metal cube which has a mass of \(3 \, \text{kg}\) and has a projected area of \(1 \, \text{m}^2\) on Earth \(90^\circ\) downward, through air at a temperature of \(25^\circ \text{C}\), what would the terminal velocity of the cube be?

All we have to do is input all the values into the formula. The acceleration due to gravity on Earth is \(9.81 \, \text{m/s}^2\). The density of air at \(25^\circ \text{C}\) is \(1.1839 \, \text{kg/m}^3\) and the drag coefficient of a cube is \(1.05\) facing downward.

The result is:
\[
V_t = 6.881101581 \, \text{m/s}.
\]

That's terminal velocity for you!

I would like to thank Mr. Tan Ping Hock and Mr. Yao Zhi Wei Adrian, my current and previous physics teachers respectively, for clearing my doubts about certain concepts within this topic of terminal velocity!

Thanks for reading!
\end{tcolorbox}

\end{document}
